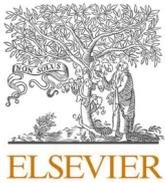
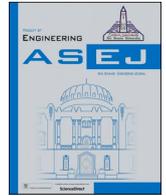

# A multi-Layer CNN-GRUSKIP model based on transformer for spatial−TEMPORAL traffic flow prediction

Karimeh Ibrahim Mohammad Ata [a,*], Mohd Khair Hassan [a,*], Ayad Ghany Ismaeel [d], Syed Abdul Rahman Al-Haddad [b], Thamer Alquthami [c], Sameer Alani [e]

[a] *Department of Electrical and Electronic Engineering, Faculty of Engineering, Universiti Putra Malaysia, 43400, UPM Serdang, Selangor Darul Ehsan, Malaysia*
[b] *Department of Computer and Communication Systems Engineering, Universiti Putra Malasia, Seri Kembangan, Selangor 43400, Malaysia*
[c] *Electrical and Computer Engineering Department, King Abdulaziz University, Jeddah 21589, Saudi Arabia*
[d] *Computer Technology Engineering, College of Engineering Technology, Al-Kitab University, Iraq*
[e] *Computer Center, University of Anbar, Iraq*

A B S T R A C T

Traffic flow prediction remains a cornerstone for intelligent transportation systems (ITS), influencing both route optimization and environmental efforts. While Recurrent Neural Networks (RNN) and traditional Convolutional Neural Networks (CNN) offer some insights into the spatial–temporal dynamics of traffic data, they're often limited when navigating sparse and extended spatial–temporal patterns. In response, the CNN-GRUSKIP model emerges as a pioneering approach. Notably, it integrates the GRU-SKIP mechanism, a hybrid model that leverages the Gate Recurrent Unit's (GRU) capabilities to process sequences with the 'SKIP' feature's ability to bypass and connect longer temporal dependencies, making it especially potent for traffic flow predictions with erratic and extended patterns. Another distinctive aspect is its non-standard 6-layer CNN, meticulously designed for in-depth spatiotemporal correlation extraction. The model comprises (1) the specialized CNN feature extraction, (2) the GRU-SKIP enhanced long-temporal module adept at capturing extended patterns, (3) a transformer module employing encoder-decoder and multi-attention mechanisms to hone prediction accuracy and trim model complexity, and (4) a bespoke prediction module. When tested against real-world datasets from California's Caltrans Performance Measurement System (PeMS), specifically PeMS districts 4 and 8, the CNN-GRUSKIP consistently outperformed established models such as ARIMA, Graph Wave Net, HA, LSTM, STGCN, and APTN. With its potent predictive prowess and adaptive architecture, the CNN-GRUSKIP model stands to redefine ITS applications, especially where nuanced traffic dynamics are in play.

## 1. Introduction

Intelligent transportation systems (ITS) play a crucial role in modern transportation engineering by enhancing road safety for drivers and pedestrians alike. The impact of traffic congestion goes beyond inconvenience, extending to higher anxiety levels, increased travel costs, and elevated air pollution. Researchers and professionals worldwide have explored solutions to tackle traffic congestion, with ITS achieving notable success in collecting, analysing, and distributing traffic data to facilitate informed decisions [1]. Traffic flow representing the mean number of vehicles on a road segment in a specific location during a defined period, can be forecasted in both short-term (10–15 min ahead) and long-term (extending to the next day) intervals. The dynamic nature of traffic conditions, influenced by factors like weather, accidents, events, public transportation, and road closures, poses a spatiotemporal challenge. These factors collectively shape the traffic environment, impacting travel times and transportation network efficiency [2]. Enhanced route planning has yielded positive outcomes in reducing traffic and air pollution [3]. Developing accurate traffic flow prediction models aims to minimize prediction errors through parameter adjustments. Predicting traffic conditions involves various methods, including parametric models (e.g., ARIMA, SARIMA, HA)[4–6] and non-parametric models (comprising machine learning models like SVR, SVM, and KNN, and deep learning models (RNN, LSTM, CNN, GCNN, GRU)[7–11]. Additionally, deep learning techniques, including attention methods, have further advanced prediction accuracy [12,13]. The capability of the parametric model-based technique (ARIMA) and its modifications to represent stochastic, seasonal, and time series that occur in the traffic flow data has led to its early acceptance in applications related to traffic flow prediction [14]. However, a few errors were noted when nonlinear data and unpredictable traffic patterns were described. Traditional machine learning methods, such as SVR, Gaussian Process, Hidden Markov Model, and K-nearest Neighbour, handle high-dimensional data and detect nonlinear correlations[15,16 17,18]. However, these approaches rely on expert-developed features and may struggle with complex spatial–temporal patterns.

Convolutional Neural Networks (CNN) can process the spatial correlation of grid-structured data, like that found in images or motion movies [19]. However, it is incorrect for the traffic graph network, further complicating the situation. Hence, Graph Convolutional Networks (GCN) are better suited for non-Euclidean spatial structures like traffic road networks [20,21]. However, GCN falls short in handling long-range dependencies and temporal patterns which calls for the integration of Recurrent Neural Networks (RNN), like LSTM, which are adept at capturing temporal relationships. This combination enhances the model's capability to handle both short-term and long-term traffic flow prediction[22]. Deep learning models have a limitation in effectively capturing complex patterns in data due to their inability to focus on relevant features. To address this, attention mechanisms are essential in traffic flow prediction. They enable models to dynamically emphasize important spatial and temporal relationships, enhancing accuracy and adaptability.

This study introduces a novel hybrid CNN-SKIPGRU architecture based on the multi-head attention transformer architecture to predict traffic flows and their long-periodic dependencies. The approach effectively combines CNN and GRU to weigh the relevance of features, leveraging spatial features from CNN, determining long-term dependencies with GRU-SKIP, and assigning weights with multi-head attention. This study presents a hybrid CNN-GRU model with a skip function proposed for accurate traffic flow prediction. Temporal and spatial data are extracted using CNN and SKIPGRU-based networks. Multi-head attention and transformer enhance relationships between projected traffic flow and other variables. The results demonstrate superior performance compared to benchmark models. The paper is structured as follows: Section 2 discusses related studies, subdivided into RNN-based models, hybrid CNN-LSTM models, and encoder-decoder models. Section 3 presents the suggested model's modules. Section 4 outlines the architecture and flowchart. Experimental results and analysis are detailed in Section 5. Lastly, Section 6 concludes and outlines future directions.

## 2. Literature review

To construct the traffic prediction model, this section methodically examines the application of the structured CNN, RNN, and transformer on spatiotemporal traffic flow data. This section explains the traffic prediction problem, the rationale for employing the structural RNN as a foundation for the traffic prediction model, how to enhance the performance of the deep learning network by adding an attention mechanism, and the drawbacks of existing models.

### 2.1. RNN based model

Recurrent neural networks (RNNs) as opposed to other deep learning structures have become more useful in recent years for processing sequential input to handle graph structure data. because the deep learning-based approach can process information more quickly and has higher generalization capabilities[23]. Predictions of traffic flows have benefited greatly from RNN-based models. For instance, LSTM is used to effectively record traffic dynamics, which are not linear. Time series prediction with extended temporal dependence is made possible by its success in resolving the issue of back-propagated error decay across memory blocks in RNN [8]. However, this makes the deep learning model more time-consuming. For this reason, the Gated Recurrent Neural network GRU is recommended, which is functionally equivalent to the Long Short-Term Memory network LSTM while significantly simplifying the model [24]. To anticipate the matrix's internal traffic flow, Dai et al. have used GRU equipped with a spatial–temporal feature selection algorithm to analyse the relevant spatial–temporal feature information. Using a GRU with spatial–temporal features outperformed using a single GRU, the results revealed [25]. The Selected Stacked Gated Recurrent Units model (SSGRU) was also proposed by Sun et al., who are leaders in the field of machine learning [26]. Thus, GRU performs similarly to LSTM. In contrast, GRU learns more quickly and with fewer parameters.

### 2.2. RNN-CNN based models

A CNN-LSTM Network was introduced by Fouladgar et al. to explore how automobiles navigate between different crossings. Both models offer n-level predictions for a range of traffic conditions (calm, medium, heavy, congested, etc.). However, this is not an issue because spatial and temporal data are controlled by distinct modules. This approach fails to adequately integrate spatial and temporal elements. To forecast traffic flow, Liu et al. proposed the Conv-LSTM model to extract the spatial–temporal parameters and combine them with the periodic features [27]. These methods are effective on networks with little or no traffic. The proposed technique, however, did not consider either the long-term recall of traffic or the relationships between different lines. Yu et al. introduced Spatiotemporal Recurrent Convolutional Networks (SRCNs) to analyse a series of static images generated from the traffic speed data and predict the network-wide traffic status to address these constraints. CNNs can capture the spatial interdependence of network-wide traffic, but Deep LSTMs learn the temporal dynamics of the traffic [28].

Zheng et al. introduced the embedding component in the LSTM-CNN model to record the categorical feature data and identify associated features. Though the CNN module can learn 2-D traffic flow data, the LSTM module can recall the earlier data indefinitely [29]. Fouladgar et al. proposed a Skip-ConvLSTM Encoder-Decoder model that uses ConvLSTM to skip spatial–temporal traffic matrix series to identify long-term time correlations. This offers insights into the missing entries of the spatial–temporal traffic matrix as a periodic time series. This page displays a series of traffic matrices for several earlier Mondays [30]. The new approach provided by Bartlett et al. differs from previous time-series predictions that employed CNNs to uncover correlations between the segments of differing temporal magnitudes. Then, with the aid of (D2), the regression model (GRU) produced a precise estimate of the traffic flow [31]. However, the challenging nature of LSTM-CNN-based models tends to increase with depth. Duan et al. recommended the use of ConvBiLSTM with an attention mechanism to extract the daily and weekly periodic characteristics. The attention mechanism may differentiate between the importance of the flow sequences at different times by using automatically assigned weights[32].

### 2.3. RNN-transformer (Encoder-Decoder)-based model

The encoder-decoder structure of deep learning has been widely applied to problems involving sequential data processing. Encoder-decoder deep learning is regarded as an all-encompassing, end-to-end solution for learning from sequence data that makes few assumptions about the sequence structure due to its remarkable performance in a variety of areas, including natural language processing tasks [33]. The qualitative analysis showed that the RNN Encoder-Decoder could effectively determine the linguistic regularities in the phrase table, which indirectly explains the quantitative improvements in the total translation performance. RNN-Encoder decoder models have thus been used to forecast traffic flow. For instance, Li et al. constructed a GRU using an encoder-decoder architecture to predict traffic patterns.

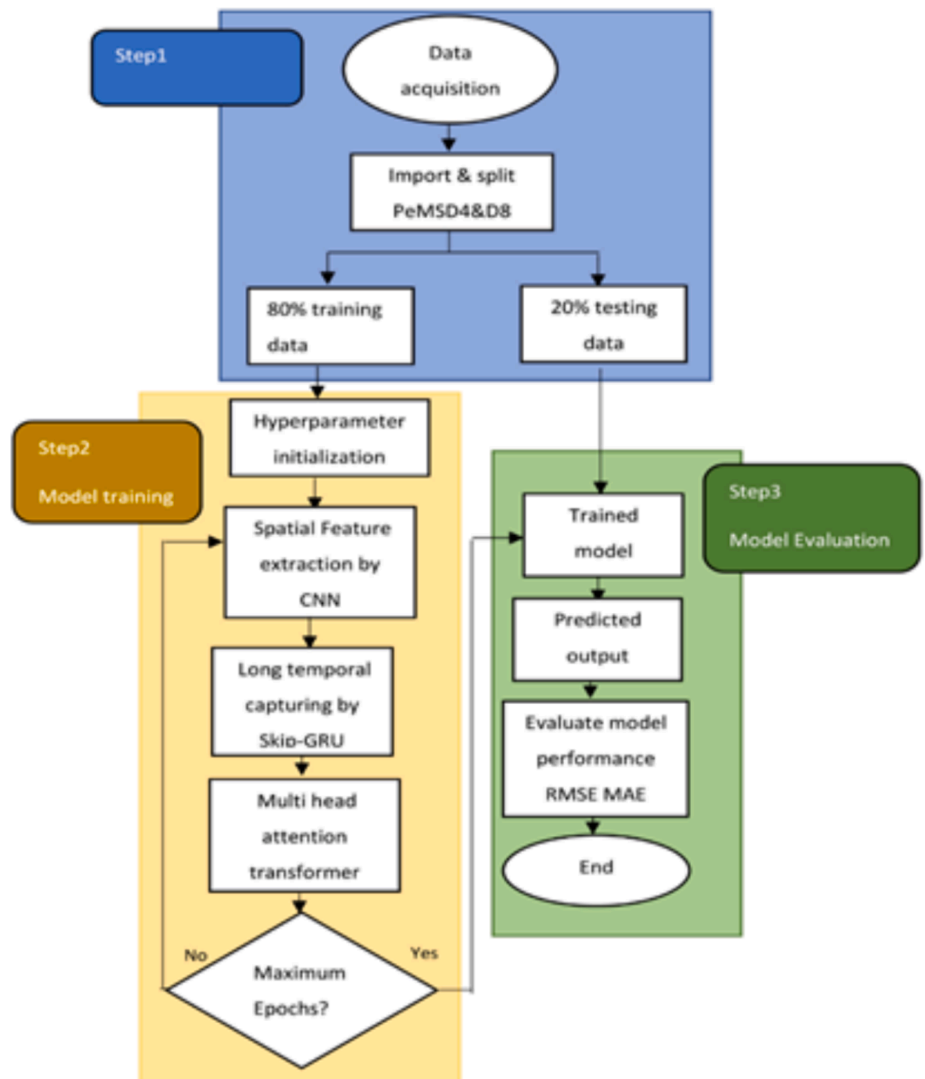

**Fig. 1.** The proposed model flowchart.

Diffusion convolution was proposed to replace matrix multiplication in GRU, resulting in the Diffusion Convolutional Gated Recurrent Unit (DCGRU) [34]. Chai et al. recommended using Multi-Graph Convolutional Networks with an LSTM-encoder-decoder model for predicting pike flow. A multi-graph convolutional operation's output sequence serves as the input for the encoder network, while the encoder's output state serves as the initial state in the decoder network [35]. An attention mechanism has been included in the RNN-encoder decoder model to increase accuracy [33]. This model uses an LSTM encoder-decoder pair with the temporal attention mechanism to forecast the forward multi-step traffic flow. He et al. proposed an LSTM-encoder-decoder and attention for both short- and long-term traffic prediction over a network. A new spatial attention model was built in the encoder to consider the significance of each connection in the network. The model learns the spatial–temporal relationships from previous traffic series. The model uses a decoder RNN outfitted with LSTM units and the temporal attention model to extract the most important and relevant historical spatial–temporal correlations from the encoder for long-term traffic prediction [36].

Wei et al. proposed an LSTM-autoencoder to extract the features of the upstream and downstream traffic flows, thereby acquiring the internal connection of traffic flow [37]. Fan et al. proposed the dual-attention architecture known as LSTM for encoders and decoders. Fan et al. proposed the dual-attention LSTM architecture for encoders and decoders. To detect the building-traffic relationships and adaptively identify the most important building sensing data, it is proposed to focus on both the temporal elements of historical connections and cross-domain attention on input data [38]. Wang et al. developed a hard attention mechanism for its application in LSTM-encoder decoders to reduce the time–space requirements of the standard attention methods [39]. Shi et al. offer an LSTM Encoder-Decoder architecture with novel skip and attention functions to analyse the long-term periodic input and encode the spatial and periodic connections [40]. The previous methods have demonstrated that RNN-based models, such as LSTM and GRU, exhibit strong capabilities in capturing temporal dependencies, CNN-based models excel in feature extraction from spatial data due to their efficient feature mapping characteristics, and the attention mechanism, especially when incorporated into the Transformer, brings a distinct advantage in handling dynamic spatial–temporal traffic flow prediction. However, the computational complexity of deep learning models, particularly LSTM-based approaches, can lead to increased processing time.

Furthermore, CNN-LSTM hybrids face challenges in e integrating spatial and temporal data, leading to potential information loss. The primary contribution of this paper lies in the integration of RNN, CNN, and transformer models within a unified framework which addresses the shortcomings of existing methods by leveraging the strengths of each model component and modifying them according to their shortcomings.

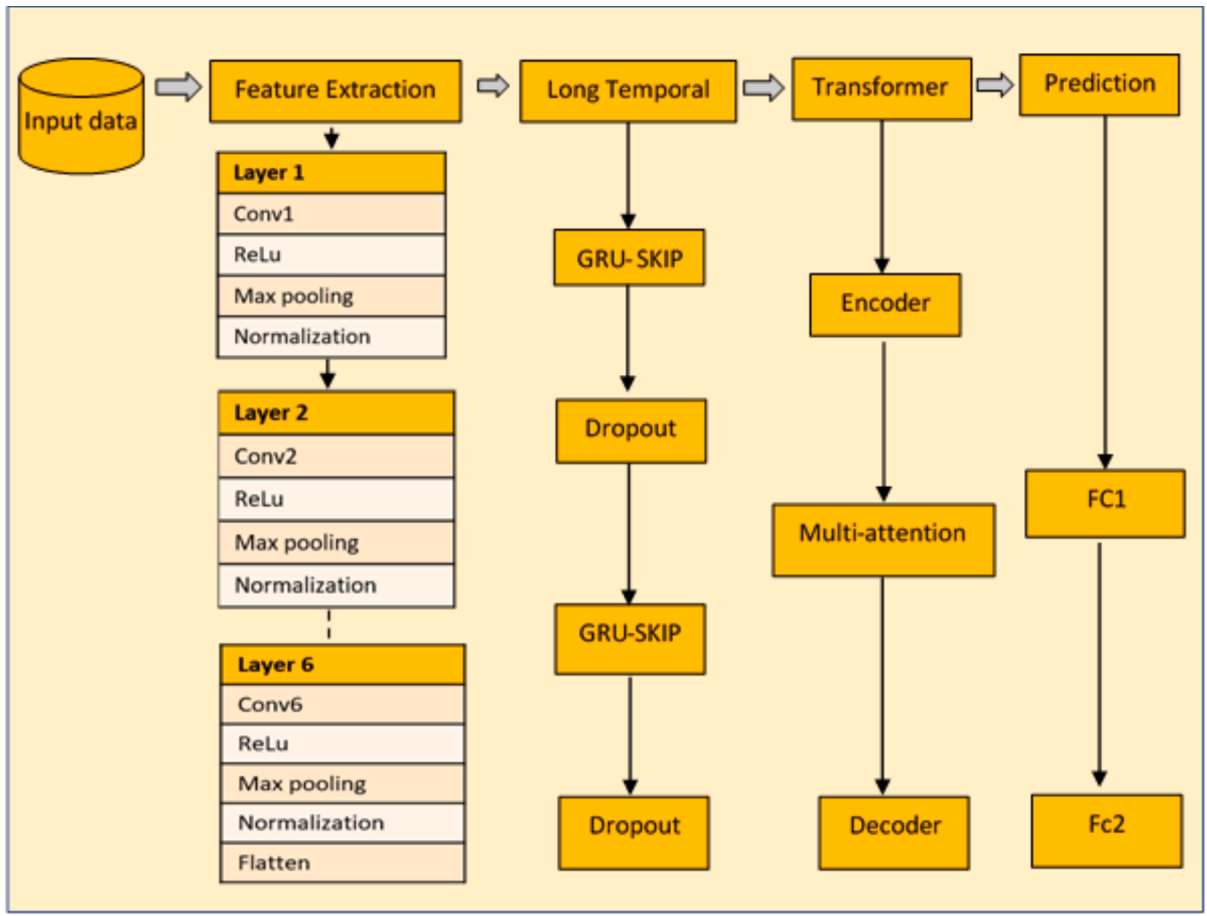

**Fig. 2.** The Architecture of the Proposed Model, where FC stands for a fully connected neural network. Bullets refer to layer3,4&5.

A significant integration of skip connections with the GRU. Skip connections facilitate the direct transfer of information across layers, which, in conjunction with the GRU, not only leverages the model's rapid learning capabilities but also addresses its vanishing gradient problem. This allows the model to retain and efficiently process longer temporal dependencies, circumventing the loss of critical long-term information seen in earlier methodologies.

Moreover, a 6-layer CNN structure has been designed to be more effective in capturing embedded spatial–temporal features. This deeper CNN allows for a more granular extraction and understanding of spatial patterns. The incorporation of transformer-based multi-head attention further bolsters the model's ability to comprehend complex spatial–temporal relationships, ensuring more accurate predictions across various time intervals and dimensions.

## 3. Methodology

In this section, a comprehensive breakdown of the proposed multilayer CNN-GRUSKIP-transformer Model. Initially, a flow chart illustrates a sequential representation of the various stages involved, from data ingestion to the final prediction output. Following this, a block diagram sheds light on the embedded modules, their connections, and the intricate mechanisms that work in tandem to predict traffic flow. This section will provide a more detailed examination of each component's role and its contribution to the model's overall efficacy.

### 3.1. The general CNN-SKIPGRU transformer flowchart process

The proposed CNN-SKIPGRU Transformer framework, as illustrated in Fig. 1, encompasses three pivotal stages: data preparation, model training, and model evaluation. During the data preparation phase, data is meticulously sourced from extant datasets. This accumulated data undergoes a rigorous standardization process and is subsequently partitioned into training and testing sets. In the model training phase, the journey commences with the determination of the initial parameter values for the nascent model. After the feature extraction procedure, the spatial attributes of the input data are systematically channelled through the layers dedicated to processing lengthy sequences. This component, adept at handling elongated sequences, harnesses the relayed spatial attributes to render a nuanced characterization of time series data spanning extensive durations. Upon the culmination of these procedures, the transformer adeptly isolates features from the output of the sequential learning segment that exhibits a striking congruence with the intended prediction results. As an iterative refinement measure, after each training epoch, parameter optimization is undertaken via the back-propagation method, persisting until the predefined epoch limit is attained. The final phase, model evaluation, involves the deployment of the calibrated model to prognosticate the outcomes based on the testing data. Employing this tripartite methodology, the presented framework is adept at predicting traffic flow patterns.

### 3.2. CNNSKIPGRU – transformer

This section presents the structure and method of the proposed traffic flow forecast approach. To extract features and increase prediction performance more effectively, a hybrid CNN, SKIPGRU, and transformer are combined into one framework. The suggested approach consists of five key modules, as shown in Fig. 2, which include input data, feature extraction, long-term dependency, multi-head attention transformer, and prediction modules. The input module processes the input data so that it can be used in the feature extraction module. To extract the spatial characteristics, the input data is initially processed through 6 CNN layers in the feature extraction module. The SKIPGRU module receives these features and uses them to extract long-term temporal data from the time slot. The SKIPGRU module eventually transmits the results to the transformer. The transformer module is utilised in multi-head attention to assign various weights to the feature input of the model, highlighting the more crucial element and assisting the model in deriving a better conclusion. To generate the final prediction, a completely connected layer was stacked as the output layer in the prediction module. The prediction error of the proposed method can be improved by selecting the optimal values for filter size, number of neurons, loss function, and kernels that constitute each of these components. The details of the proposed model are presented in Table 1.

**Table 1**
Details of proposed modules.

| Module | layer | Number of neurons |
|---|---|---|
| **Feature extraction module** | Conv1 | 16*5*1 |
| | Max-pooling | 16–02-2001 |
| | Conv2 | 32*3*1 |
| | Max-pooling | 32/2/1 |
| | Conv3 | 64*3*1 |
| | Max-pooling | 64/2/1 |
| | Con4 | 128*3*1 |
| | Max-pooling | 128/2/1 |
| | Conv5 | 128*3*1 |
| | Max-pooling | 128/2/1 |
| | Conv6 | 128*3*1 |
| | Max-pooling | 128/2/1 |
| | flatten | – |
| **Long-term dependency module** | GRU-SKIP1 | 128 |
| | GRU-SKIP2 | 64 |
| **Transformer Module** | 6-layers Encoder | 32 |
| | 3-layers multi-head attention | |
| | 6-layers decoder | |
| **Prediction block** | FC1 | 32 |
| | FC2 | 1 |

The proposed technique containing the building modules has been described below:

1. Input module

Because of the precise traffic flow model, the temporal data in the historical day should be detected in the input module. As a result, the suggested model considers the present day as short-term, while the long-term regular dependency represents the entire time frame.

2. Feature extraction module

The Feature Extraction Module is an integral component of the architecture, comprising a one-dimensional, 6-layer Convolutional Neural Network (CNN) meticulously designed to discern localized traffic flow patterns and features within distinct road structures as shown in Fig. 3. This module is composed of 6 convolutional layers that strategically uncover intricate details of the traffic flow dynamics. Each of these convolutional layers is coupled with a subsequent pooling layer, where the pooling operation facilitates the abstraction of spatial information to a higher level. This process accentuates meaningful features while concurrently mitigating extraneous noise, ultimately contributing to a refined representation. The architecture incorporates a deliberate augmentation of filter size for the initial convolutional layer, distinguishing it from subsequent layers. This design choice aims to adeptly capture more comprehensive spatial characteristics in the data, fostering the recognition of broader road layouts. Furthermore, the module integrates the power of multiple convolutional and pooling layers arranged in tandem.

Furthermore, the stacking mechanism allows the extraction of pro-gressively higher-level features from the input, thereby enriching the overall perceptibility of the input data. Temporal correlations ingrained within the input data are effectively harnessed through the dynamic sliding filters employed within the convolutional layers. Notably, the module navigates the intricacies of kernel sizes, recognizing that smaller kernels could be less proficient in suppressing high-frequency signals, especially when confronted with the module's sophisticated architecture. As a convolutional layer advances through to the Max-pooling layer, a rectified linear unit (ReLU) activation function is engaged. This strategic choice serves to both avert the potential impediments of gradient vanishing and explosion and expedite the model's convergence. Addressing stability and performance optimization, a pivotal regularizing strategy, known as the normalization strategy, is adeptly applied after each convolutional layer. This strategy not only enhances the network's training performance but also substantially reduces the likelihood of internal covariate shift as shown in Fig. 4. The training process is accelerated by using the layer-by-layer feature normalisation method. Before being normalised to the standard distribution, the features of each layer are controlled to the ideal distributions. Batch Normalisation changes the signal at each concealed layer in the following manner:

$$\mu = \frac{1}{n}\sum_i Z^{(i)} \quad (2)$$

$$\sigma^2 = \frac{1}{n}\sum_i |(Z^{(i)} - \mu)_0^2 \quad (3)$$

$$Z_{norm}^{(i)} = \frac{Z^{(i)} - \mu}{\sqrt{\sigma^2 - \epsilon}} \quad (4)$$

$$\widehat{Z} = \gamma * Z_{norm}^{(i)} + \beta \quad (5)$$

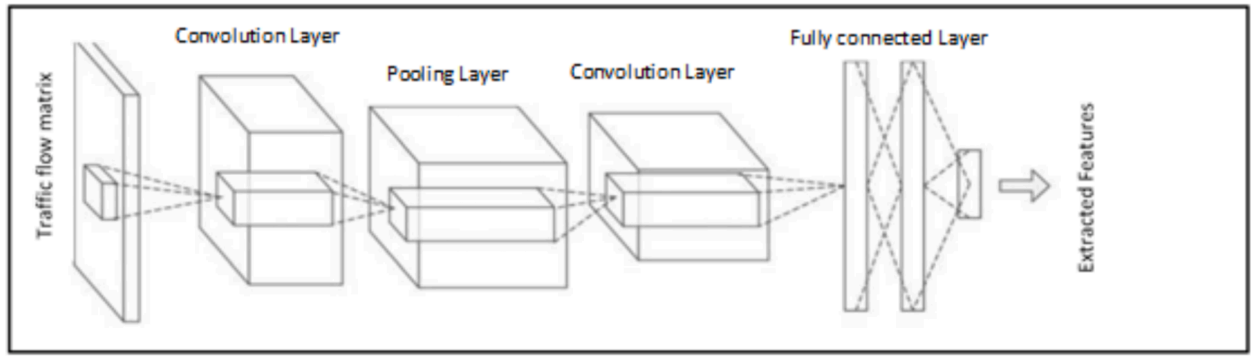

**Fig. 3.** CNN structure [44].

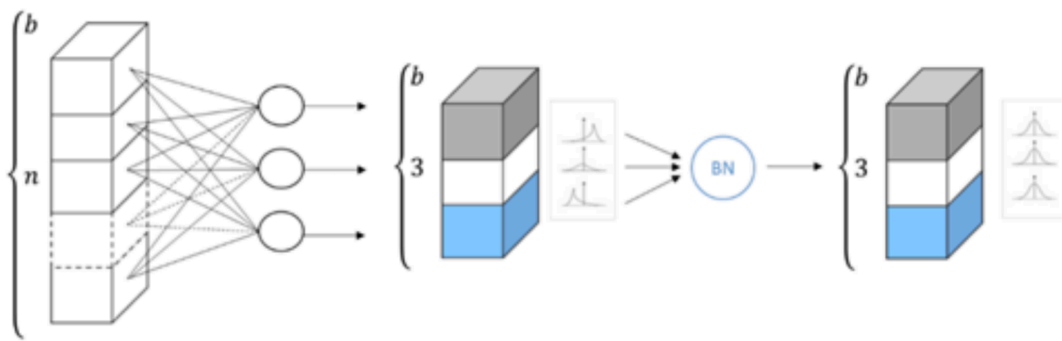

**Fig. 4.** Normalization approach where BN stands for batch normalization.

Here $n$ denotes the batch size, $Z^{(i)}$ and $\widehat{Z}$ indicate the input and output observation values in the batch. The average value in the batch sample is indicated by $\mu$. The standard deviation in the batch sample is equivalent to $\gamma$. $\epsilon$ indicates a constant which maintains the numerical stability as near zero, $\beta$ indicates a bias parameter, while $\gamma$ is the scaling parameter. For ensuring that none of the features were lost during the convolution operations, the padding type is maintained. The flattening layer is added to the final layer in the CNN to flatten the multidimensional output into the single-dimensional data as the following GRU-SKIP module needs 1-D input data.

The efficiency in extracting refined features from traffic data is significantly enhanced by employing the CNN Training Algorithm with Batch Normalization, as outlined in Algorithm 1.

**Algorithim1**: CNN Training with Batch Normalization for Traffic Data Feature Extraction

| |
|---|
| **Input**: Traffic data X (i, j, t) for lanes i, positions j, and timestamps t = 0, 1, …, n-1 |
| **Output**: Refined feature set F |
| **Configuration**: |
| Size of each data batch: 64 |
| Total training iterations: 500 |
| // **Initialize the CNN** |
| 1: CNN←Initialize CNN () |
| // **Training Loop** |
| 2: for epoch ← 1 to Number of epochs do |
| // **Divide data into batches** |
| 3: **for** each batch in X divided into chunks of size Batch size do |
| 4:     **for** each layer L in CNN **do** |
| 5:         **if** L is Convolutional then |
| 6:             L. output ← Convolution (L. input) |
| 7:             mu ← Mean (L. output) |
| 8:             sigma_sq. ← Variance (L. output, mu) |
| 9:             L. output_norm ← Normalize (L. output, mu, sigma_sq.) |
| 10:            L. output_activated ← ReLU (L. output_norm) |
| 11:        **else if** L is Max Pooling, then |
| 12:            L. output ← Max Pooling (L. input) |
| 13:        **end if** |

(*continued on next column*)

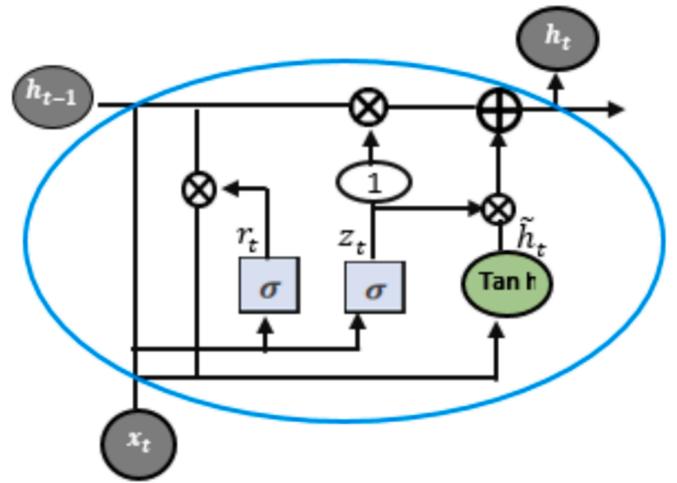

**Fig. 5.** GRU cell [8].

(*continued*)

**Algorithim1**: CNN Training with Batch Normalization for Traffic Data Feature Extraction

| |
|---|
| 14:        **end for** |
| 15:        F←Flatten (CNN. output) |
| 16:        Update the CNN parameters with backpropagation |
| 17:    **end for** |
| 18:    **return F** |
| 19: **End** |

## 4. Long-term sequence module

The Long-Term Sequence Module significantly captures the temporal patterns inherent in the properties extracted through the Feature Extraction Module. Moreover, it facilitates the identification of supplementary features derived from prolonged dependencies over time. The main component of this module is the GRU model which plays an important role in capturing intricate long-term temporal patterns within the data. Its unique architecture, featuring gating mechanisms, empowers it to selectively retain and update information over extended sequences as shown in Fig. 5. At each time step, the GRU takes in two pieces of information: the current traffic data input denoted as $x_t$ and the hidden state from the previous time step $h_{t-1}$, which contains information from the past. The GRU calculates the updated gate $z_t$ using the following equation:

$$z_t = \sigma(W_z \cdot [h_{t-1}, x_t] + b_z) \quad (6)$$

This gate determines the extent to which the previous hidden state $h_{t-1}$ should be carried over to the next state. The update gate is a value between 0 and 1, obtained by passing the linear combination of $h_{t-1}$ and $x_t$ through a sigmoid function $\sigma$. A value closer to 1 indicates greater retention of past information, while a value closer to 0 suggests discarding more of this information. Simultaneously, the GRU computes the reset gate $r_t$ using the following equation:

$$r_t = \sigma(W_r \cdot [h_{t-1}, x_t] + b_r) \quad (7)$$

This gate decides how much of the past information should be forgotten or reset when computing the new memory content. Essentially, it allows the GRU to adaptively forget or remember information based on the current input and past state. Next, the GRU forms the candidate's hidden state $\widetilde{h}_t$ using the following equation:

$$\widetilde{h}_t = \tanh(W \cdot [r_t * h_{t-1}, x_t] + b) \quad (8)$$

This state represents the new memory content that the network proposes to store in the current hidden state, blending new input data $x_t$ with relevant past information modulated by the reset gate $r_t$. The candidate state combines the past information with the current input to generate a potential new state. Finally, the GRU updates its hidden state $h_t$ for the current time step using the following:

$$h_t = (1 - z_t)*h_{t-1} + z_t*\widetilde{h}_t \qquad (9)$$

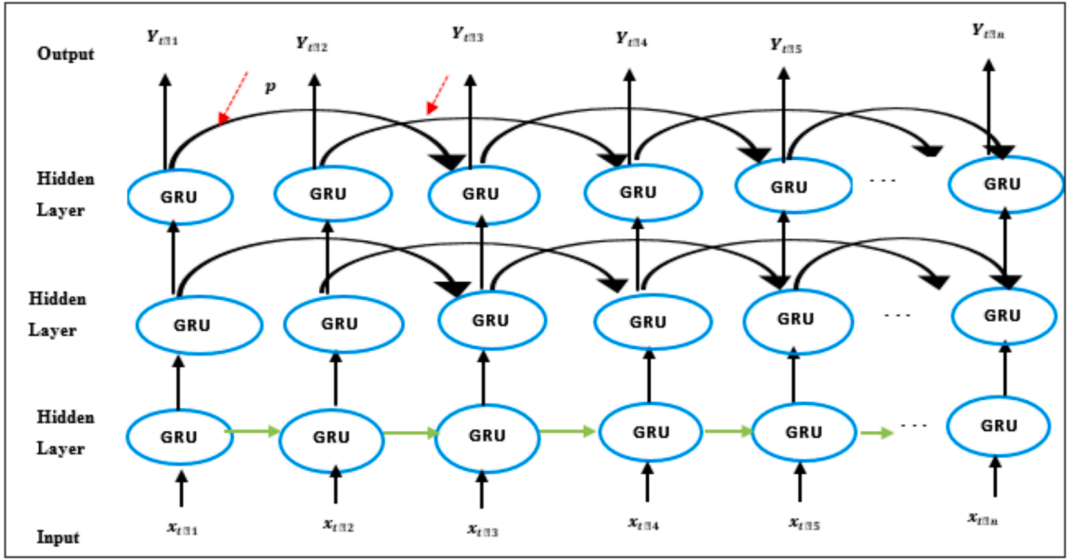

**Fig. 6.** GRU with skip connections.

This equation balances the old state $h_{t-1}$ and the new candidate state $\widetilde{h}_t$ based on the update gate $z_t$. It determines the final output for the current time step, which is critical for making predictions about traffic flow at this moment. The output hidden state $h_t$ becomes the input for the next time step, ensuring a continuous flow of information through time. GRU updates its state at each time step, considering both new inputs and historical information. This allows it to make informed predictions about future traffic patterns by effectively balancing the influence of recent and past traffic data. However, GRU in this way suffers from the challenge of handling very long-term dependencies effectively. Due to the nature of its architecture, the GRU might struggle with the vanishing gradient problem, where gradients become so small over many time steps that the model stops learning effectively from data points that are far back in the sequence. This can be particularly limiting in traffic flow prediction scenarios, where certain patterns or trends may span across extended periods.

Hence, GRU has been enhanced by skip connections to provide a direct pathway for the gradient to flow through, mitigating the risk of gradient vanishing. This is especially beneficial for learning long-term dependencies in traffic data, ensuring that critical historical information is not lost over time. Fig. 6 shows GRU layers with skip connections that jump two units in each layer. In a standard GRU, the hidden state for the current time step $h_t$ is calculated as a combination of the previous hidden state $h_{t-1}$ and the current candidate's hidden state $\widetilde{h}_t$. However, with skip connections, the hidden state can also directly incorporate information from a further past state, say $h_{t-j}$, where $j$ is the skip step. The modified equation for the GRU with a skip connection can be represented as:

$$h_t = (1 - z_t)*h_{t-j} + z_t*\widetilde{h}_t \qquad (10)$$

In this revised formulation, $h_{t-j}$ refers to the hidden state from a 'skipped' past time step. By directly using $h_{t-j}$ in the calculation of $h_t$, the model more effectively retains and utilizes information from more distant past states. This approach is advantageous in scenarios where traffic flow patterns show dependencies or cycles over extended time intervals. The input data pertinent to this module captures the period length, $j$, of the time slot. This temporal attribute is crucial in distinguishing and identifying the sequence to be skipped, enhancing the model's ability to capture the intricacies of temporal dependencies. The skip sequence complexity aids in the comprehensive understanding of the dataset's temporal dynamics as follows:

$$p = \left\{y_{T-j\times n}, y_{T-j\times (n-1)}, \cdots, y_{T-j}\right\} \text{ where } j \times n \leq T \qquad (11)$$

For a target series $y = \{y_1, y_2, \cdots, y_{T-1}\}$, each skip sequence $p$ is utilized by the GRU for extracting the periodic trend. The GRU unit's operation, with the skip connection, includes the concatenation of the hidden state $h_{t-j}$ at time $t-j$ and the input $x_t$ at time , where $\sigma$ denotes the sigmoid function and * indicates the operation of dot multiplication. The periodic characteristic value $Y$ at time $T$ is then determined as:

$$y\_skip_T = W_j h_T + b_z \qquad (12)$$

where $b_z$ and $h_T$ refer to a bias term and weight of matrix in a linear layer. After every GRU SKIP layer, the dropout strategy is used to prevent the problem of data over-fitting. Dropout refers to the fact that only a section of the network's neurons is selected randomly and trained, in comparison to all the neurons, to improve the model's performance using periodic time series datasets. In essence, a few of the neurons in every iteration train stop receiving any output and become dormant. It promotes the network to seek more beneficial traits and improves the generalisability of the model. In summary, the integration of skip connections into the GRU model addresses the vanishing gradient issue and enhances the model's capability to learn and remember longer sequences of data. This results in improved prediction accuracy, especially for traffic flow patterns characterized by long-term dependencies as demonstrated in algorithm 2. Fig. 6 shows the integration of GRU with skip connections.

**Algorithm 2:** Integration of Long-Term Sequence Learning with GRU-Skip Connections

**Input:** Feature set F from Feature Extraction Module, of time slot j
**Output:** Predicted traffic pattern y_skip_T
1: Initialize GRU with skip connections
2: Initialize hidden state to zero for t = 0
3: h_t ← 0
// **Process each feature set in F sequentially through the GRU**
3: **For** each x_t in F **do**
4:            z_t ← CalculateUpdateGate(x_t, h_t) using Eq. (6)
5:            r_t ← CalculateResetGate(x_t, h_t) using Eq. (7)

*(continued on next page)*

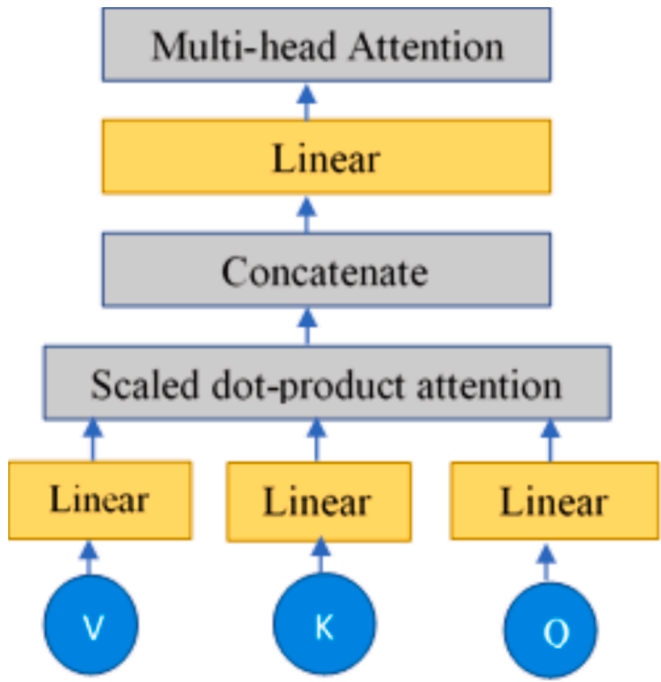

**Fig. 7.** multi-head attention layers.

(*continued*)

| Algorithm 2: Integration of Long-Term Sequence Learning with GRU-Skip Connections |
|---|
| 6:       h_tilde_t ← CalculateCandidateState(x_t, h_t, r_t) using Eq. (8) |
| 7:       h_t ← UpdateHiddenState(h_t, h_tilde_t, z_t) using Eq. (10) |
| 8:       h_t ← ApplyDropout(h_t) |
| 9: **End for** |
| // **Generate predictions using the final hidden state** |
| 10: y_skip_T <- W_j * h_T+b_z // Predict using Eq. (12) |
| 11: **Return** y_skip_T |
| 12: **End** |

## 5. The transformer module

The introduction of the Transformer Module, as detailed in Algorithm 3, marks a significant advancement in the proposed model. This module utilizes a sophisticated structure consisting of six encoder and six decoder layers, integrated with a multi-head attention mechanism. This layer empowers the model to simultaneously focus on data from various representation subspaces at distinct positions, thereby highlighting crucial variables that influence prediction outcomes.

$$\text{Multi-Head}(Q, K, V) = \text{Concat}(\text{head}_1, \cdots, \text{head}_h)W^O \quad (11)$$

where $Head_i = \text{Attention}(QW_i^Q, KW_i^K, VW_i^V)$. Here, the projections refer to parameter matrices $W_i^Q \in \mathbb{R}^{d_{model} \times d_k}, W_i^K \in \mathbb{R}^{d_{model} \times d_k}, W_i^V \in \mathbb{R}^{d_{model} \times d_v}$ and $W^O \in \mathbb{R}^{hd_v \times d_{model}}$. this study presents, h = 8 parallel attention layers, or heads. For each of these, $d_k = d_v = d_{model}/h = 64$. The entire computational cost is equivalent to one-head full-dimensional attention since each head's dimension has been lowered. Multi-head attention parallel layers are shown in Fig. 7. This transformational process encompasses both the Encoder and Decoder Stacks. The encoder encompasses N=6 layers, each with two sublayers. Layer 1 incorporates a multi-head self-attention mechanism, while Layer 2 integrates a fully connected feed-forward network with positional completeness. Residual connections link these sublayers post-normalization. The output of a sublayer is computed using the formula Layer Norm (x + "Sublayer"(x)), emphasizing its function. The decoder consists of N=6 stacked identical layers. It augments each encoder layer's output with a 3rd sub-layer for multi-head attention. Like the encoder, each sublayer is normalized and enveloped by residual connections. Each location transforms a fully linked feed-forward network, with attention sub-layers embedded in both the encoder and decoder. An efficient transformation driven by ReLU activation bridges these two linear transformations. The embeddings and SoftMax form the next component. Learned embeddings convert input and output tokens into dimension vectors, following earlier sequence transduction models. The decoder employs these strategies to translate its output into projected next-token probabilities, coupled with traditional learned linear transformation and SoftMax functions. A recommended approach employs a shared weight matrix between the two embedding layers and a linear pre-SoftMax transformation. Positional Encoding constitutes the final facet. Both embeddings and positional encodings share the same dimensional model, encompassing both learned and hardwired positional encodings. This intricate ensemble of components in the Transformer module crafts a multifaceted approach to interpreting spatial–temporal patterns, fundamentally enhancing the model's prediction accuracy within the context of traffic dynamics. Fig. 8 shows the transformer components.

| Algorithm 3: Transformer Module for Traffic Prediction |
|---|
| **Input:** Sequence of encoded features F from Long-Term Sequence Module |
| **Output:** Predicted output sequence Y with enhanced feature representations |
| 1: Initialize Transformer Model with 6 encoder layers, 6 decoder layers, and 3 attention heads |
| // **Apply 6-layer Encoder Stack** |
| 2:       **for** i = 1 to Number of Encoder Layers **do** |
| 3:       F←Apply Multiheaded Self Attention (F, Numb Attention Heads) |
| 4:       F←Apply Position Wise Feed Forward(F) |
| 5:       F←Apply Layer Norm and Residual(F) |
| 6:       **End for** |
| 7:       D←Prepare Decoder Input(F) |
| // **Apply 6-layer Decoder Stack** |
| 8:       **for** i = 1 to Number of Decoder Layers do |
| 9:       D←Apply Multi Headed Self Attention (D, Numb Attention Heads) |
| 10:      D←Apply Multi Headed Encoder Decoder Attention (D, F, Numb Attention Heads) |
| 11:      D←Apply Position Wise Feed Forward(D) |
| 12:      D←Apply Layer Norm and Residual(D) |
| 13:      Y←Convert to Output Sequence(D) |
| 14: end for |
| 15: **Return** Y |
| 16: **End** |

## 6. Prediction module

Prediction Module for Traffic Flow Forecasting is a critical component in the overarching architecture of our traffic prediction model. This module takes the processed feature set H, meticulously derived from the Transformer Module, and transforms it into actionable traffic forecasts. The heart of this algorithm lies in its two-layered structure: a fully connected layer followed by an output layer. The fully connected layer, equipped with an activation function like ReLU or tanh, plays a vital role in introducing non-linearity to the model. This non-linearity is crucial as it enables the model to capture complex patterns and relationships within the traffic data, which linear models might overlook. The output layer, applying the SoftMax function, translates the processed features into final traffic forecast predictions (P). This SoftMax layer is particularly important for classification tasks, as it helps in determining the probability distribution over different traffic states, allowing for a more nuanced and detailed prediction. The integration of these layers in Algorithm 4 not only enhances the model's ability to make accurate predictions but also ensures that these predictions are interpretable and relevant for practical traffic management applications. This algorithm, therefore, stands as a testament to the sophistication and effectiveness of modern traffic flow forecasting systems.

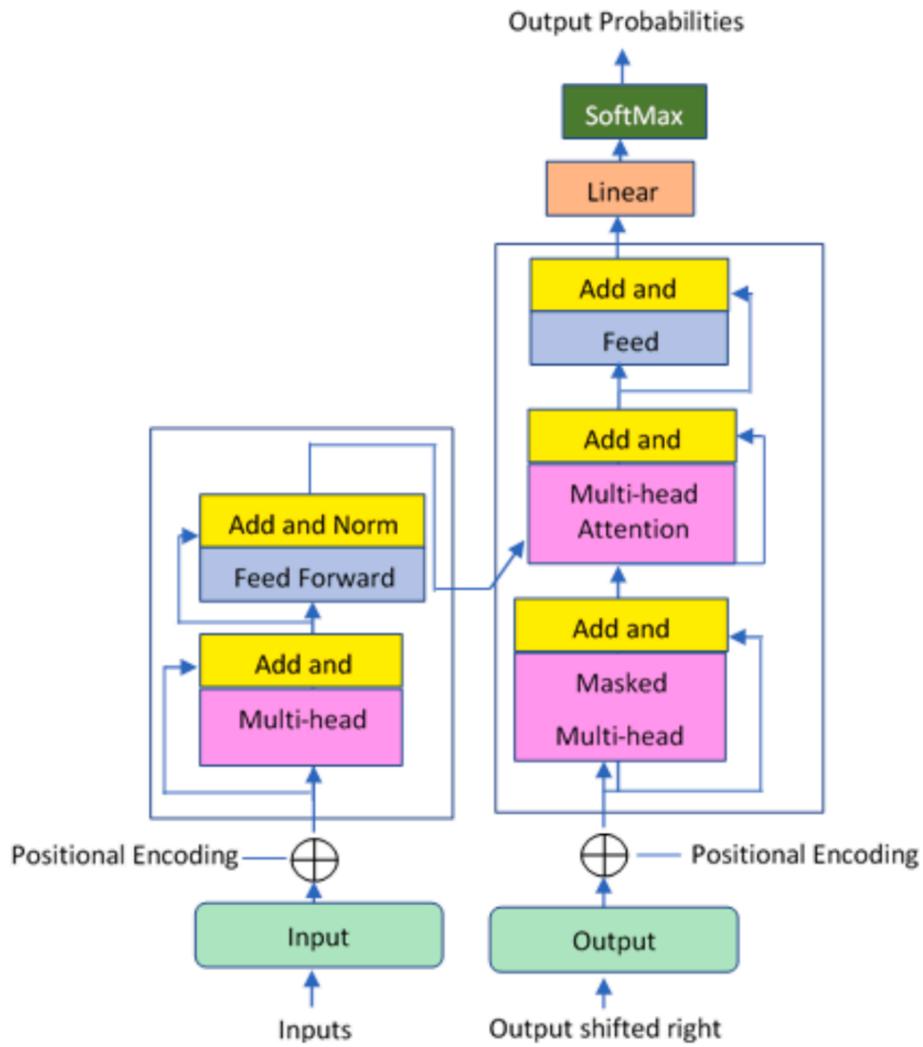

**Fig. 8.** The Transformer − model architecture.

| Algorithm 4: Prediction Module for Traffic Flow Forecasting |
| --- |
| 1:    **Input:** Processed feature set H from Transformer Module |
| 2:    **Output:** Traffic forecast predictions P |
| 3:    **Procedure** Prediction Module |
| 4:       Initialize the weights for the FC layer W_fc and output layer W_out |
|    // **Apply the fully connected layer with nonlinearity** |
| 5:       H' <- Activation Function (W_fc * H+b_fc) |
| 6:       P <- SoftMax (W_out * H' + b_out) // Apply the output layer with SoftMax for classification |
| (*continued on next column*) |

| Algorithm 4: Prediction Module for Traffic Flow Forecasting |
| --- |
| 7:       **Return P** |
| 8:       **End** Procedure |
| 9:       Function Activation Function(x) |
|    // **Choose an activation function such as ReLU or tanh** |
| 10:        Return activated_x |
| 11:       **End Function** |
| 12:        Function SoftMax(x) |
|    // **Apply the SoftMax function for classification problems** |
| (*continued on next page*) |

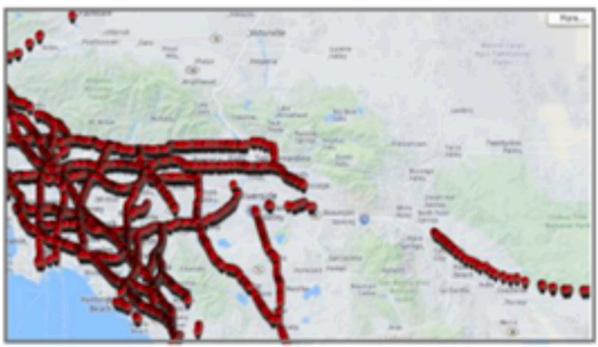 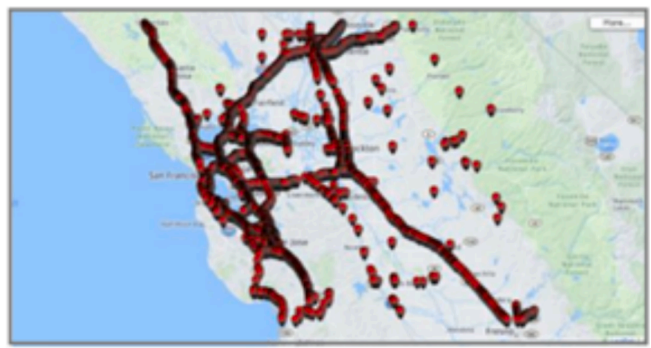

**Fig. 9.** Distribution of sensors in district 4 and district 8.

**Table 2**
Distribution of data framework.

|  | Count | Mean | std | MIN | 25 % | 50 % | 75 % | MAX |
|---|---|---|---|---|---|---|---|---|
| Index | 3,035,520 | 1,517,759.50 | 876,279.29 | 0.00 | 758,879.75 | 1,517,759.50 | 2,276,639.25 | 3,035,519.00 |
| Sensor ID | 3,035,520 | 84.50 | 49.07 | 0.00 | 42.00 | 84.50 | 127.00 | 169.00 |
| flow | 3,035,520 | 230.68 | 146.22 | 0.00 | 110.00 | 215.00 | 334.00 | 1,147.00 |
| occupancy | 3,035,520 | 0.07 | 0.05 | 0.00 | 0.04 | 0.06 | 0.08 | 0.90 |
| speed | 3,035,520 | 63.76 | 6.65 | 3.00 | 62.60 | 64.90 | 67.40 | 82.30 |
| Hour | 3,035,520 | 11.50 | 6.92 | 0.00 | 5.75 | 11.50 | 17.25 | 23.00 |
| Day of week | 3,035,520 | 4.00 | 2.02 | 1.00 | 2.00 | 4.00 | 6.00 | 7.00 |
| month | 3,035,520 | 7.50 | 0.50 | 7.00 | 7.00 | 7.50 | 8.00 | 8.00 |
| Minutes | 3,035,520 | 27.50 | 17.26 | 0.00 | 13.75 | 27.50 | 41.25 | 55.00 |
| Origin ID | 3,017,664 | 80.74 | 48.80 | 1.00 | 37.00 | 79.00 | 121.00 | 168.00 |
| Destination ID | 3,017,664 | 83.83 | 49.84 | 0.00 | 40.00 | 82.00 | 127.00 | 169.00 |
| Cost | 3,017,664 | 343.11 | 90.20 | 6.30 | 320.90 | 350.20 | 379.60 | 816.20 |
| Day | 3,035,520 | 16.00 | 8.94 | 1.00 | 8.00 | 16.00 | 24.00 | 31.00 |
| Weekend | 3,035,520 | 0.29 | 0.45 | 0.00 | 0.00 | 0.00 | 1.00 | 1.00 |
| Week | 3,035,520 | 30.50 | 2.59 | 26.00 | 28.00 | 30.50 | 33.00 | 35.00 |

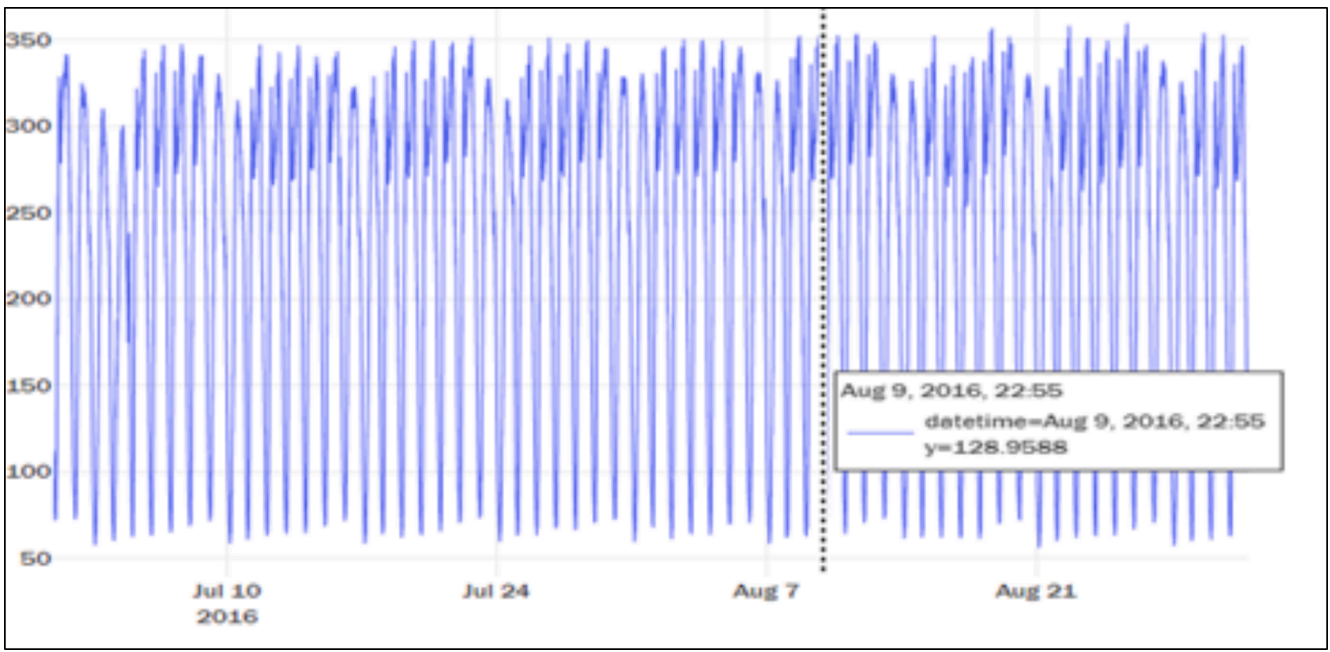

**Fig. 10.** Average congestion in district 8.

| (*continued*) | |
|---|---|
| **Algorithm 4: Prediction Module for Traffic Flow Forecasting** | |
| 11: | Return SoftMax_x |
| 12: | **End Function** |

## 7. Experiment and results

### 7.1. Datasets

PeMS refers to a real-world traffic dataset developed by the California Department of Transportation that is employed to assess the proposed technique. PeMS continually collects data from the loop detectors every 30 s. Caltrans is divided into 12 districts that together contribute to the daily data volume of 2 GB. This data collection is frequently employed as a benchmark for traffic predictions [41].

Among the numerous important metrics gathered by the detector nodes include the average lane occupancy, average vehicle speed, and total flow. Two different types of PeMS data sets were utilised for this purpose, as shown in Fig. 9.

- PeMS District 4 (PeMSD4): Data was collected from 307 sensors placed along 29 roads in the San Francisco Bay Area. The data collection includes information from January and February of 2018. The data is aggregated every 5 min, with a total of 288 records each day (24 h).
- PeMS District 8 (PeMSD8): In San Bernardino, 170 detectors were positioned along 8 highways from July through August 2016 and data were collected every 5 min.

The dataset comprises 16 columns and 3,035,520 rows. The mean, standard deviation, and shape of data distribution are summarised using descriptive statistics, which are displayed in Table 2. The number of observations without NA or null values (Count), maximum and lowest possible values (Maximum and Minimum, respectively), the average value (Mean), and standard deviation (Std) are presented in the appropriate columns. The proposed framework also accounts for geographical and temporal information.

The geographical data includes both the sensor sites and distances between them, whilst the temporal data is divided into 5-min intervals and consists of row data with 3 properties (Speed, Flow, and occupancy). Additionally, utilising PeMS data, the general traffic conditions in

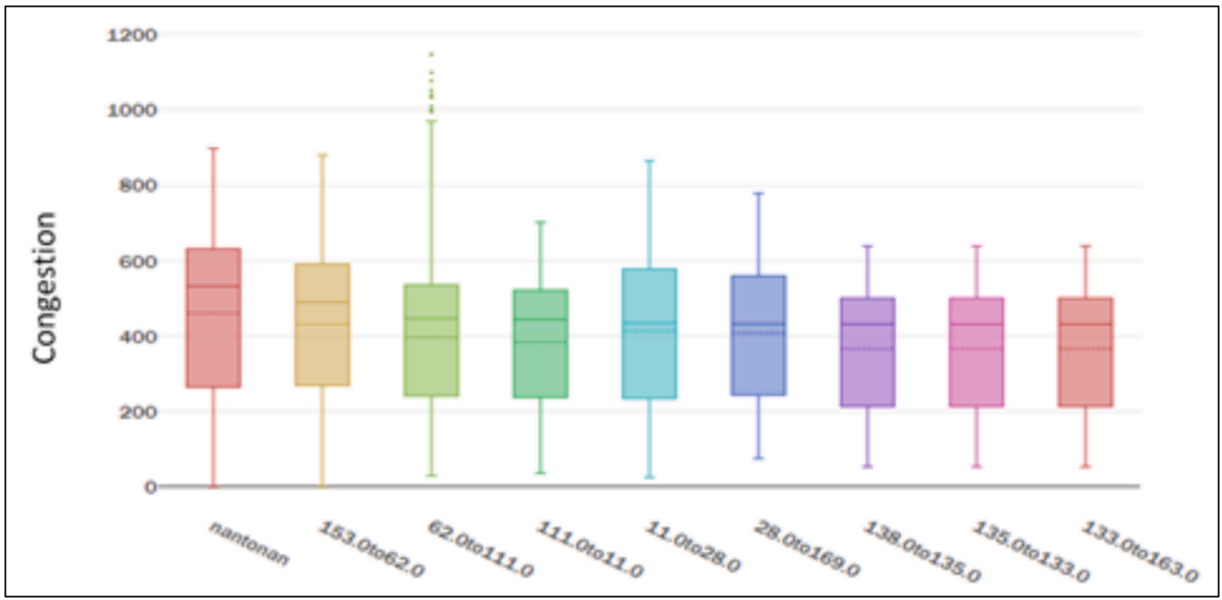

**Fig. 11.** Distribution of congestion by direction.

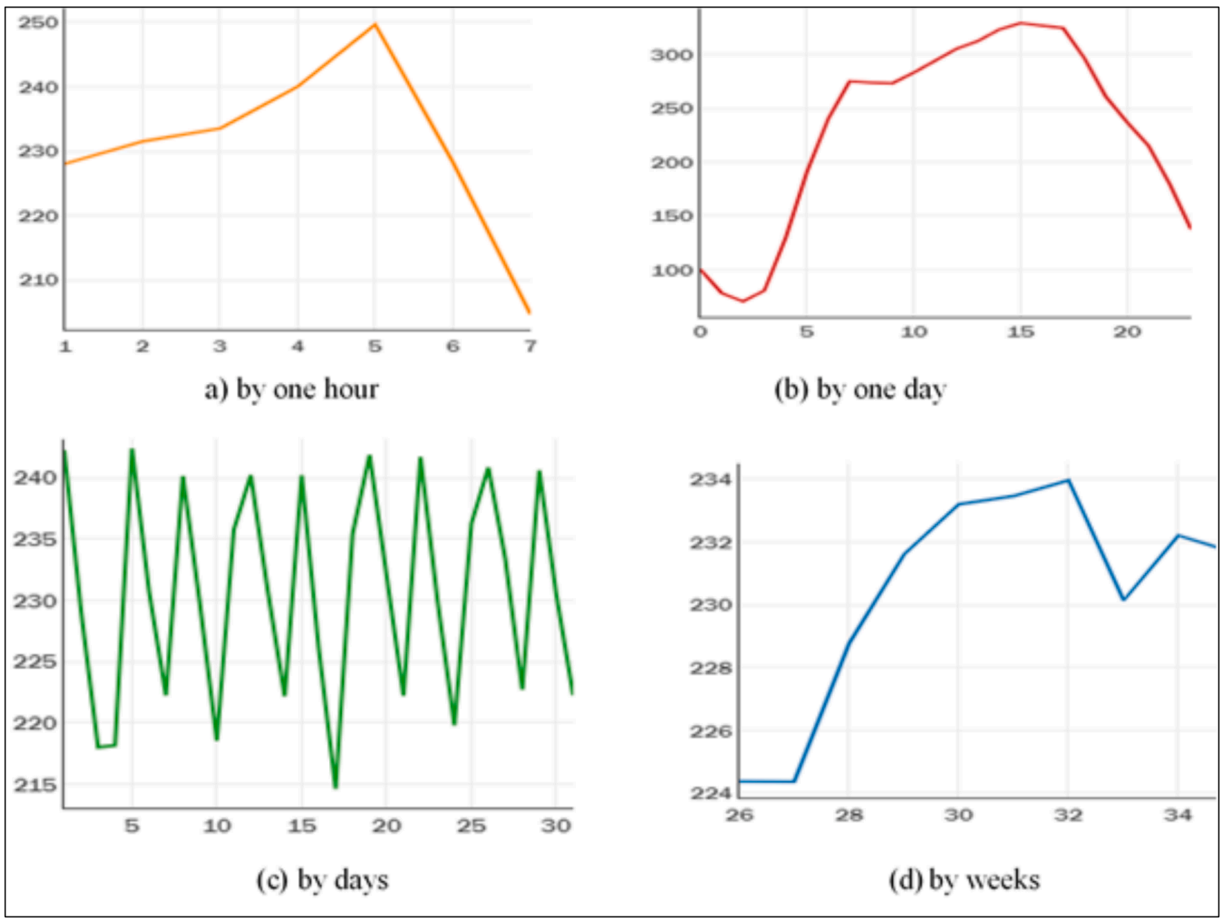

**Fig. 12.** Ddistribution of vehicles on a specific road by time.

California were analysed. Fig. 10 depicts the congestion in District 8. Density values are displayed along the Y-axis, and congestion is displayed along the X-axis. It displays the typical degree of congestion in July and August 2016. A weekly repetitive pattern can be displayed by plotting time on the X-axis and the congestion levels are plotted on the Y-axis. The intensity of congestion changes with the direction of movement, as seen in Fig. 11. Moreover, Fig. 12 depicts a scatter plot of the distribution of vehicles on a specific road by week, day, and hour. Here, 80 % of the data was employed as the training set, 10 % was used as the verification set, and the remaining 10 % was used as the test set,

Table 3

Comparison of Traffic Prediction Model Performances on the PeMSD4 Dataset.

| PeMSD4 | | | | | | | | |
|---|---|---|---|---|---|---|---|---|
| | Time | ARIMA | HA | LSTM | Graph Wave Net | STGCN | APTN | CNN-SKIPGRU |
| RMSE | 10 | 58.05 | 54.14 | 45.72 | 37.09 | 38.16 | 31 | 29.03 |
| | 20 | 60 | 57.2 | 46 | 38.61 | 39 | 31.6 | 29.03 |
| | 30 | 60.56 | 57.9 | 46.2 | 39.81 | 39.3 | 31.67 | 29.04 |
| | 40 | 61.98 | 58.4 | 46.53 | 39.82 | 39.6 | 31.7 | 29.04 |
| | 50 | 62.3 | 60.1 | 46.7 | 40.99 | 39.89 | 31.86 | 30 |
| | 60 | 63.02 | 64.07 | 46.9 | 43.1 | 41.5 | 40.7 | 30.07 |
| MAE | 10 | 35.19 | 36.7 | 30 | 23.14 | 27.02 | 19.15 | 16.2 |
| | 20 | 37.17 | 38 | 30.2 | 23.7 | 27.98 | 19.9 | 16.4 |
| | 30 | 40.9 | 39.8 | 30.8 | 24.8 | 28.1 | 19.98 | 16.48 |
| | 40 | 41.7 | 43 | 30.8 | 24.7 | 28.19 | 20.3 | 17.62 |
| | 50 | 43.16 | 44.52 | 31 | 24.2 | 29.4 | 20.74 | 17.8 |
| | 60 | 45 | 46.3 | 31.11 | 25.83 | 30.02 | 21 | 17.98 |

all in chronological sequence.

### 7.2. Implementation

The success of a trained model dramatically depends on the hyper-parameters chosen for it. Most of the hyper-parameter adjustments were made by trial and error. There is a comparison of many optimization techniques, such as stochastic gradient descent (SGD) [41], Momentum [42], and Adam [43]. The validation set performance is most significant with SGD and Adam, as shown by the comparative results. Weights and biases are updated by backpropagation using a mean square error loss function (MSE, RMSE, MASE). The dropout was set to = [0.0,0.1,0.2,0.3, 0.4,0.5] and the learning rate to = 0.001. This promotes a more consistent pace throughout the educational process. The batch size is 64, and the epoch number is 500. This study uses h = 8 heads or parallel attention layers. Additionally, the size of the feature representation (vis) and the dimension of all GRU units' hidden state (m) is set to 128. The proposed experiment uses a 64-bit batch size.

### 7.3. Evaluation metrics

Three metrics root mean square error, mean absolute error, and mean absolute scaled error were employed in this study to assess the accuracy of the new proposed model and compare it to the existing models. The below-mentioned equations have been used for determining the maximum absolute standard error, the mean squared error, and the root mean squared error:

$$RMSE = \sqrt{\frac{1}{n}\sum_{i=1}^{n}(y_i - \widehat{y}_i)^2} \quad (12)$$

$$MAE = \frac{1}{n}\sum_{i=1}^{n}|y_i - \widehat{y}_i| \quad (13)$$

$$RMSE = \sqrt{\frac{1}{\Omega n}\sum_{i=1}^{\Omega}\sum_{k=1}^{n}|y_i - \widehat{y}_i|^2} \quad (14)$$

In the above equations, n refers to the no. of predicted data; $\widehat{y}_i$ is the predictive value, while $y_i$ is the actual value. As quality indicators for regression issues, standardized measures of analysis error (RMSE, MAE, and MASE) are frequently used. The accuracy of predictions can be increased by reducing the root-mean-squared error, mean absolute error, and mean-squared error, which are used as metrics of model performance. These estimates consider the values ranging from 0 and Infinity, where 0 indicates the optimal performance.

Table 4

Comparison of Traffic Prediction Model Performances on the PeMSD8 Dataset.

| PeMSD8 | | | | | | | | |
|---|---|---|---|---|---|---|---|---|
| RMSE | 10 | 43.23 | 44.03 | 36.98 | 27.87 | 30.32 | 24.74 | 22.82 |
| | 20 | 54.76 | 46.81 | 37.2 | 30.5 | 30.63 | 25.8 | 23.6 |
| | 30 | 56.3 | 47.93 | 37.11 | 30.99 | 30.74 | 25.87 | 23.98 |
| | 40 | 58.1 | 48.02 | 37.17 | 31.82 | 30.85 | 26.76 | 23.99 |
| | 50 | 59.4 | 50.7 | 37.5 | 31.97 | 31.3 | 26.9 | 24.3 |
| | 60 | 60.1 | 52 | 37.8 | 32.1 | 31 | 27.8 | 24.86 |
| MAE | 10 | 24.2 | 29.52 | 23.1 | 17.54 | 20.7 | 15.62 | 12.3 |
| | 20 | 26.07 | 30.9 | 23.98 | 17.9 | 21 | 16.8 | 12.9 |
| | 30 | 29.95 | 34.65 | 23.65 | 18.42 | 21.72 | 16.7 | 13 |
| | 40 | 30.98 | 37.54 | 23.32 | 18.57 | 22.51 | 16.99 | 13.8 |
| | 50 | 32.8 | 39.62 | 22.8 | 19.92 | 22.91 | 18 | 13.98 |
| | 60 | 38.12 | 40.52 | 22.92 | 20 | 23.87 | 18.4 | 14.21 |

### 7.4. Comparison between the proposed model and other competitive models

The suggested CNN-SKIPGRU transformer model was compared to the below benchmark models.

1. HA: Historical Average (HA) analyses the seasonal fluctuations in tourist demand and generates appropriate forecasts. The projection period is 1 week, and it is based on the average amount of data derived within the same period in the earlier weeks.
2. ARIMA [42]: This model, which combines moving average and autoregressive elements, presents a generalization of the ARMA model.
3. LSTM: Long-short-term memory network is a variation of the RNN model that may detect long-term time correlation [33].
4. Graph Wave Net [43]: A graph convolution network can find latent graph patterns in the data owing to an in-built adjacency matrix. It presents a CNN-based graph convolution layer that may maintain latent spatial relationships while learning an adaptive adjacency matrix from the input.
5. STGCN [20]: By merging graph convolutional layers with convolutional sequence learning layers, the Spatial-Temporal Graph Convolution Network (STGCN) simulates the spatial and temporal connections.
6. APTN [44]: A Spatial-Temporal Attention Approach for Traffic Prediction describes an end-to-end traffic forecasting system that takes into account spatial–temporal periodic dependency.

### 7.5. Analysis of the results

The suggested model was compared to 6 baseline models on PEMSD4 and PEMSD8 for 10, 20, 30, 40, 50, and 60 min. The error rate determined by the RMSE and MAE evaluation metrics over the time ranges is shown in Table 3 and Table 4. In terms of the assessment measures, the proposed CNN-GRUSKIP achieves the best performance using two datasets. Traditional time-series prediction methods (HA and ARIMA) perform poorly because they only consider historical data and neglect the spatial components. Furthermore, the worst results are generated by the forecasts with a 60-minute prediction time. A deep learning method, called LSTM, ignores spatial relationships and just concentrates on the temporal data sequence. Deep learning models such as Graph Wave Net, STGCN, and APTN showed a better performance than LSTM when used with traffic networks because they can consider temporal and spatial connections. The experiment, therefore, illustrates the necessity of taking the traffic network topology into account with time series data when trying to determine traffic flows.

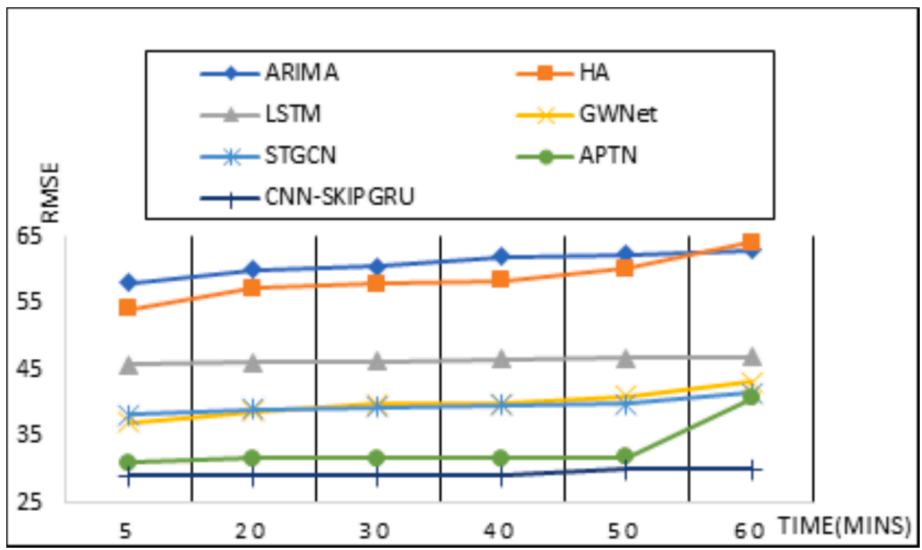

**Fig. 13.** RMSE for PeMSD4.

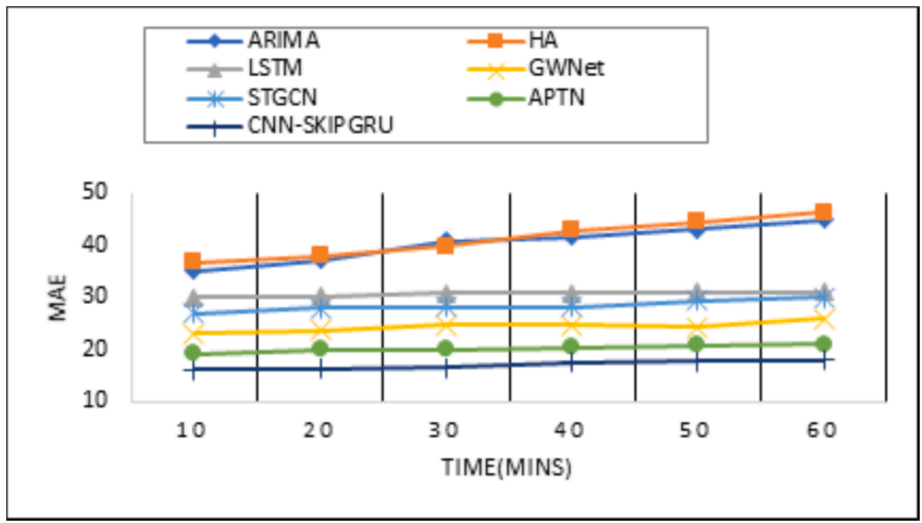

**Fig. 14.** MAE for PeMSD4.

The proposed model, CNN-GRUSKIP, performs significantly better than the state-of-the-art technique, as it achieves the lowest RMSE and MAE values across all 6 prediction intervals. The outcomes show that the proposed model offers a more accurate analysis of the spatial–temporal characteristics of traffic data. Moreover, the proposed model architecture, which cleverly integrates convolutional neural networks with skip connections and gated recurrent units, is well-suited to capturing the complex spatial and temporal dependencies characteristic of traffic data. Such proficiency is crucial for reliable predictions is displayed in Fig. 13 and Fig. 14 for PeMSD4, as well as Fig. 15 and Fig. 16 for PeMSD8.

While traditional models like ARIMA struggle with the non-linear patterns present in traffic flow, especially over extended forecast horizons, more sophisticated models such as LSTM, STGCN, GWNet, and APTN have been developed to address these complexities. However, even these advanced models fall short of the performance benchmarks set by CNN-SKIPGRU. STGCN and APTN, despite being specifically designed to handle spatial–temporal data, do not quite match the proposed model's results. STGCN, a model known for its ability to capture spatial dependencies through graph convolution, and APTN, which leverages attention mechanisms, show improved results over ARIMA but still underperform compared to CNN-SKIPGRU, as evidenced in the figures. This disparity underscores CNN-SKIPGRU's robustness and its adept handling of traffic forecasting intricacies. The comparison between PeMSD4 and PeMSD8 datasets offers further insights into the models' capabilities. PeMSD8, with its smaller dataset size, typically poses a greater challenge to predictive models due to the reduced amount of data available for training. Nevertheless, CNN-SKIPGRU maintains the highest level of accuracy on both datasets, indicating its ability to derive effective data representations even when the data volume is limited. This adaptability is a notable strength in practical

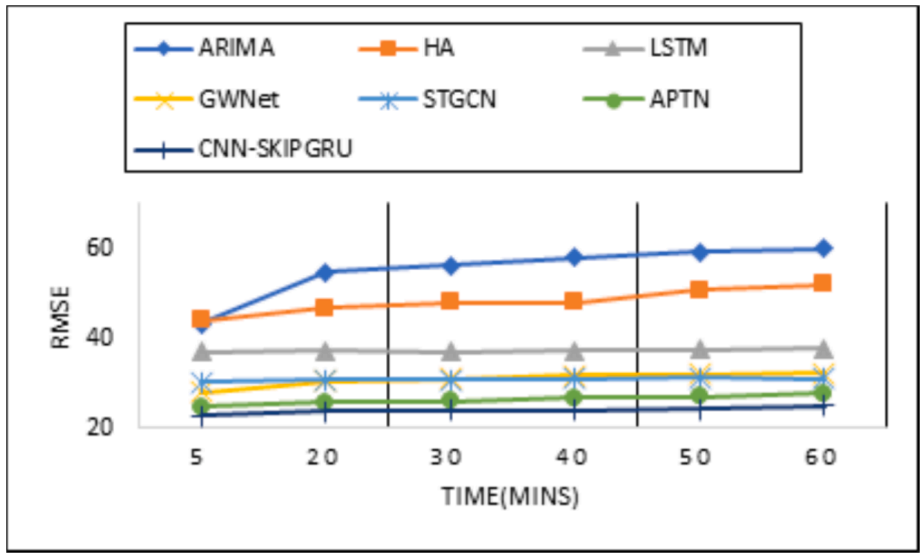

**Fig. 15.** MAE for PemSD8.

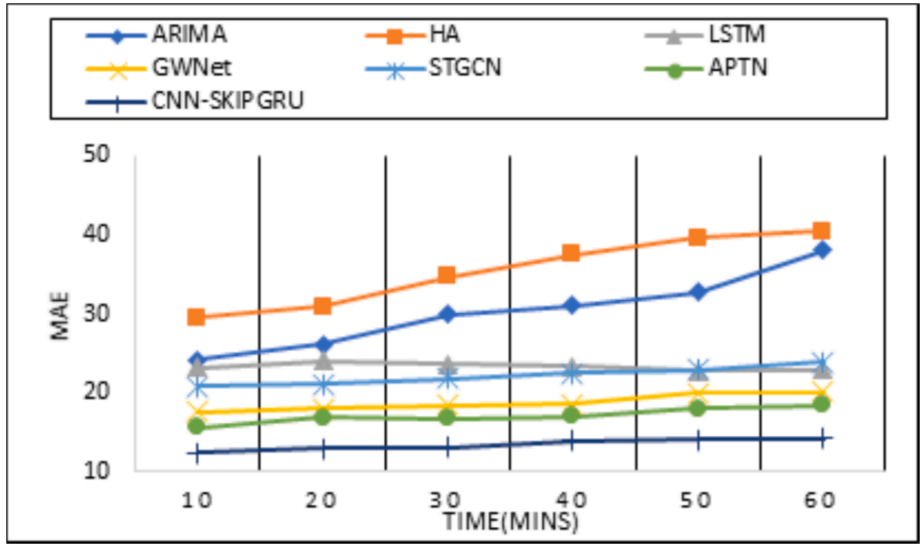

**Fig. 16.** RMSE for PeMSD8.

applications where data collection may be constrained.

Moreover, the consistent performance of CNN-SKIPGRU across both datasets suggests a level of robustness and generalizability that is less pronounced in models like HA and ARIMA, and even in more advanced models like STGCN and APTN. These models exhibit a more significant performance degradation with the smaller PeMSD8 dataset, a limitation not shared by CNN-SKIPGRU. Additionally, a comparison of execution time per epoch for various traffic prediction models on the PeMSD4 and PeMSD8 datasets has been done as shown in Fig. 17. The execution times are comparable across both datasets, suggesting that the complexity of the models, rather than the size of the data, predominantly affects computational efficiency. The proposed CNN-SKIPGRU model shows competitive execution times when compared to the other models. It does not have the shortest execution time, which suggests that while it offers superior prediction accuracy, as discussed earlier, it does so at the cost of computational efficiency. This trade-off is often observed in more complex models that integrate multiple components to capture spatial–temporal relationships more effectively. The traditional models such as ARIMA and HA, which typically have less computational complexity, seem to have shorter execution times. This aligns with expectations since these models have simpler structures and require fewer calculations per epoch. Advanced models like LSTM, Graph Wave Net, STGCN, and APTN exhibit varying execution times. Models like Graph Wave Net and STGCN, designed to capture complex spatial–temporal dependencies, might have longer execution times due to their intricate architectures involving graph convolutions and attention mechanisms. In conclusion, the CNN-SKIPGRU model distinguishes itself by surpassing both traditional and several advanced baseline models on the PeMSD4 and PeMSD8 datasets. Its robust performance across datasets of varying sizes and complexities solidifies its potential as a versatile and reliable asset in the domain of traffic management and forecasting. While the model does not boast the lowest execution time, this aspect is mitigated by its superior accuracy, which is paramount for non-real-time applications. For real-time traffic forecasting, where latency is more critical, the trade-off between the model's execution time and its predictive performance must be carefully considered. Nonetheless, the CNN-SKIPGRU model's ability to deliver enhanced predictions may well justify the longer execution times in scenarios where forecast precision is crucial, potentially contributing to significant advancements in traffic forecasting and management practices.

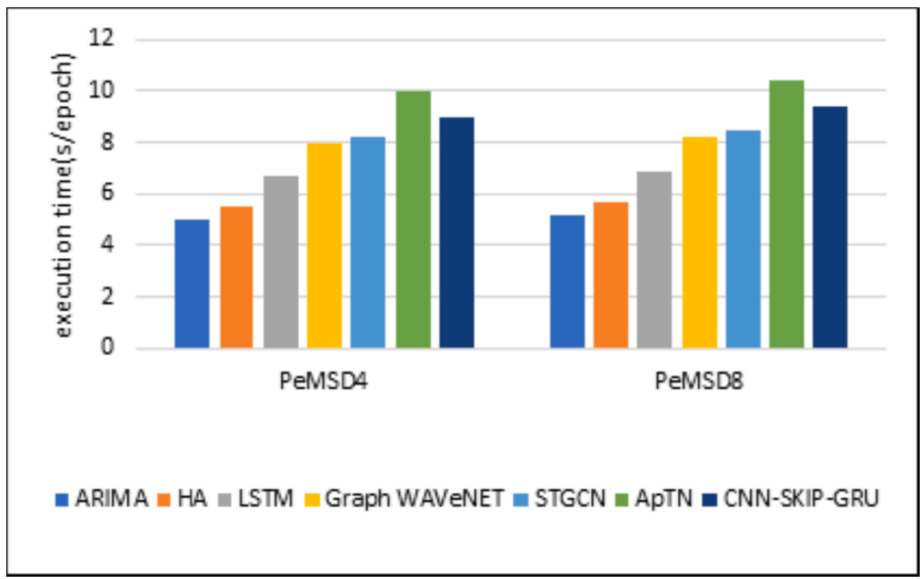

**Fig. 17.** Comparison of the time consumption on PeMSD4 and PeMSD8.

**Table 5**
Performance of model components on PeMSD4.

| PeMSD4 | | |
|---|---|---|
| Model variants | RMSE | MAE |
| CNN-GRU/F | 31.8 | 18.3 |
| CNN-GRUSKIP/T | 29.98 | 16.54 |
| CNN-GRUSKIP/P | 30.53 | 17.67 |
| CNN-GRUSKIP | 29.3 | 16.2 |

**Table 6**
Performance of model components on PeMSD8.

| PeMSD8 | | |
|---|---|---|
| Model variants | RMSE | MAE |
| CNN-GRU/F | 28.12 | 17.43 |
| CNN-GRUSKIP/T | 26.8 | 14.5 |
| CNN-GRUSKIP/P | 23 | 12.9 |
| CNN-GRUSKIP | 22.82 | 12.3 |

*7.6. Ablation analysis*

The proposed CNN-SKIPGRU model includes 4 different components, such as feature extraction module, temporal module, correlation module, and prediction module. The CNN-SKIPGRU and its different versions have been compared to further investigate the efficacy of every component, in the following manner:

1. CNN-SKIPGRU/F: Eliminate the feature extraction module.
2. CNN-KIPGRU/T: Eliminate the periodical input. When the long-temporal module component is removed, the transformer's input only accepts the weighted input.
3. CNN-SKIPGRU/P: Eliminate the prediction module. The results predicted by the proposed model are the transformer's output. Tables 5 and 6 present the results of each module. It was noted that the feature extraction module was the most effective component of this system. The RMSE showed an increase from 29.3 to 31.08 without the need for any spatial awareness.

The Prediction Module is ranked as the Second Most Influential in Data. This demonstrates how fully linked layers can be used to improve the output of the neural network. Since long-temporal dependency modelling is a feature of traffic data, it is a minor component of the proposed CNN-GRUSKIP model. Fig. 18 indicates and proof that without a temporal module, the RMSE increases from 29.3 to 29.98.

**8. Conclusions**

This study offers a general CNN-SKIPGRU-transformer strategy and an efficient module for long-term spatial–temporal dependence in predicting traffic flow since traffic flow data is complex and nonlinear. The input data were adjusted to account for different periods before it was fed into the model, to ensure that the model could extract important data from the data sequences. Then, a 6-layer CNN is used to extract the traffic's dependence on the locations of spatial sensors. SKIPGRU includes the spatially extracted traffic features to address the long-term pattern dependency. The accuracy of the generated predictions in the multi-head attention and transformer is most heavily influenced by the SKIPGRU output features. To acquire the desired prediction, the highest-weight features are included in the fully connected layers. The proposed multi-layer model shows a better performance than several benchmark models in simulation, such as deep learning, conventional machine learning, and time series analysis methods. The suggested model uses a lightweight attention mechanism that consumes less memory, CNN parallelization, and persistent gradients from GRU. However, since this method needs multiple layers to record the long-term temporal dependence, some crucial data would inevitably be lost. Further studies would include a few extraneous factors such as weather, points of attraction, and social activities to improve the accuracy of the estimates.

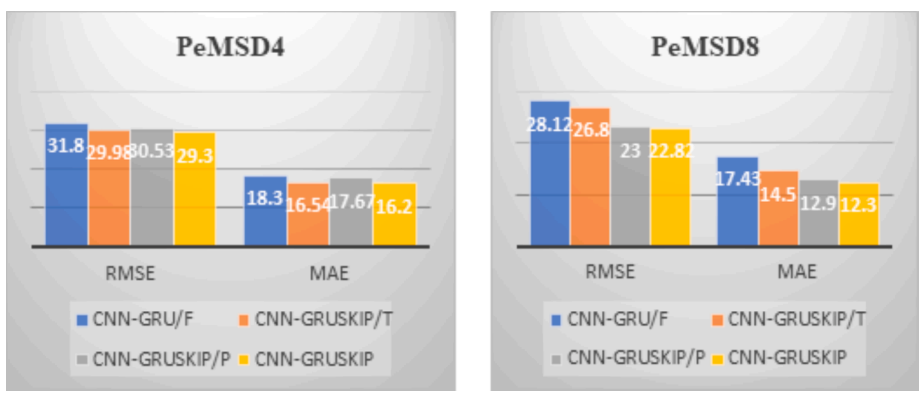

**Fig. 18.** Performance of Model Variants on The PeMS Dataset.

**Declaration of Competing Interest**

The authors declare that they have no known competing financial interests or personal relationships that could have appeared to influence the work reported in this paper.

**Acknowledgement**

This work is supported by a research grant from the Research, Development, and Innovation Authority (RDIA), Saudi Arabia, grant no. 13010-Tabuk-2023-UT-R-3-1-SE.

fx1 **Karimeh Ibrahim Mohammad Ata** is an Artificial Intelligence Engineer based in Riyadh, Saudi Arabia. She has a strong background in artificial intelligence engineering, with a focus on deep learning, big data analysis, and algorithm development. She is currently a Ph.D. candidate at UPM in Malaysia, where her research focuses on traffic flow prediction using deep learning models such as LSTM and CNN. She also completed a Master of Science in Computer and Embedded System Engineering from UPM, where she worked on solving optimization problems in smart parking systems. she holds a bachelor's degree in Computer Engineering from FBSU in Saudi Arabia and has published several research papers in shortest path algorithms, Particle swarm optimisation algorithm, and prediction by deep learning .

fx2 **Prof. Madya Ir. Dr. Mohd Khair Bin Hassan** is currently the Deputy Dean (Research & Innovation) at the Faculty of Engineering. He holds a Ph.D. in Automotive Engineering from UPM, Serdang, Malaysia, with expertise in control systems. With a strong educational background, including a master's degree in electrical engineering and a Bachelor's degree in Electrical and Electronic Engineering, he has extensive experience as a lecturer and senior lecturer at UPM. Dr. Mohd Khair's skills include system identification, modeling, MATLAB Simulink, artificial intelligence, and model-based design and calibration. He has been involved in projects related to vehicle electrical loads, hybrid artificial intelligence, and control system development.

fx3**Ayad Ghany Ismaeel**, Professor of Computer Science with M.Sc. degree in Applied Computer from Computer National Center CNC/ Ministry of Higher Education and Scientific Research-IRAQ and a Ph.D. degree in Computer Science (Computer) was obtained from Technology University Baghdad-IRAQ. His solid understanding in qualification and his administration skills in context, passion for learning, and consultative leadership style are the driving forces behind a progressively successful 32+ year career.

fx4**Professor Dr. Syed Abdul Rahman Al-Haddad Syed Mohamed** is a highly accomplished academic in the field of Computer and Communication Systems Engineering. He is currently a Professor at the Department of Computer and Communication Systems Engineering, Faculty of Engineering, Universiti Putra Malaysia (UPM). With a Ph.D. in Electrical, Electronics, and System Engineering, he specializes in areas such as speech recognition, natural language understanding, and image processing. Professor Syed Abdul Rahman has published extensively in reputable journals and conferences and holds memberships in IEEE and MySET. His research focuses on topics like video segmentation, polynomial design, and image noise removal.

fx5**Professor Dr.Thamer Alquthami** is a competent energy strategist and organizational change leader with over 17 years of experience in the government, higher education, and energy industry. They have expertise in electric power, renewable energy, and energy planning, and have led major projects and developed energy policies and regulations. With a Ph.D. in Power System Operation & Renewable Energy, they have supervised students, conducted research, and delivered advanced training in the field. They have a strong publication record and hold memberships in professional organizations. As a resilient and adaptable leader, they thrive in fast-paced environments and excel in making analytical decisions.

fx6**Sameer Alani** was born in Iraq in 1989. He received a B.S. degree in computer engineering and an M.Sc. degree in wireless communication and computer networking technology from the National University of Malaysia (UKM) in 2017. He received his PhD degree from UTEM University in 2022. His research interests include antenna applications, wireless communication, and networking technology.